\documentclass[sigconf, nonacm]{acmart}

\usepackage{adjustbox}
\usepackage{amsmath}
\usepackage{bbm}
\usepackage{xcolor}
\usepackage{url}
\usepackage{caption}
\usepackage[subrefformat=parens,labelformat=parens]{subcaption}
\usepackage[capitalise]{cleveref}
\usepackage{booktabs}

\usepackage{tikz}
\usetikzlibrary{positioning, arrows, arrows.meta, fit, calc}
\tikzset{>={Latex[scale=1.25]}}

\AtBeginDocument{%
  \providecommand\BibTeX{{%
    \normalfont B\kern-0.5em{\scshape i\kern-0.25em b}\kern-0.8em\TeX}}}

\setcopyright{acmcopyright}
\copyrightyear{2022}
\acmYear{2022}
\acmDOI{XXXXXXX.XXXXXXX}

\acmConference[SIGSPATIAL '22]{30th ACM SIGSPATIAL International Conference on Advances in Geographic Information Systems}{November 1--4, 2022}{Seattle, WA}
\acmPrice{15.00}
\acmISBN{978-1-4503-XXXX-X/18/06}

\definecolor{left} {HTML}{222222}

\definecolor{t1} {HTML}{F642C9}
\definecolor{t2} {HTML}{F6C942}
\definecolor{t3} {HTML}{42F66F}
\definecolor{t4} {HTML}{426FF6}

\definecolor{lightblue}{RGB}{3, 100, 252}
\definecolor{orange}{RGB}{252, 186, 3}
\definecolor{pink}{RGB}{255, 181, 240}

\pgfdeclarehorizontalshading{delta_shade}{4cm}{
color(0cm)=(white);
color(.2cm)=(white);
color(.25cm)=(lightblue); 
color(.31cm)=(white);
color(.5cm)=(white);
color(.55cm)=(lightblue); 
color(.61cm)=(white);
color(.9cm)=(white);
color(.95cm)=(lightblue); 
color(1.1cm)=(white);
color(1.2cm)=(white);
color(1.3cm)=(lightblue); 
color(1.5cm)=(white)
}

\begin{document}

\title{Meta-Learning over Time for Destination Prediction Tasks}

\author{Mark Tenzer}
\email{mtenzer@novateur.ai}
\affiliation{%
  \institution{Novateur Research Solutions}
  \streetaddress{20110 Ashbrook Place, Suite 275}
  \city{Ashburn}
  \state{Virginia}
  \country{USA}
  \postcode{20147}
}

\author{Zeeshan Rasheed}
\email{zrasheed@novateur.ai}
\affiliation{%
  \institution{Novateur Research Solutions}
  \streetaddress{20110 Ashbrook Place, Suite 275}
  \city{Ashburn}
  \state{Virginia}
  \country{USA}
  \postcode{20147}
}

\author{Khurram Shafique}
\email{kshafique@novateur.ai}
\affiliation{%
  \institution{Novateur Research Solutions}
  \streetaddress{20110 Ashbrook Place, Suite 275}
  \city{Ashburn}
  \state{Virginia}
  \country{USA}
  \postcode{20147}
}

\author{Nuno Vasconcelos}
\email{nuno@ucsd.edu}
\affiliation{%
  \institution{UC San Diego}
  \streetaddress{9500 Gilman Drive}
  \city{La Jolla}
  \state{California}
  \country{USA}
  \postcode{92093}
}

\renewcommand{\shortauthors}{Tenzer, et al.}

\begin{abstract}
  A need to understand and predict vehicles' behavior underlies both public and private goals in the transportation domain, including urban planning and management, ride-sharing services, and intelligent transportation systems. Individuals' preferences and intended destinations vary throughout the day, week, and year: for example, bars are most popular in the evenings, and beaches are most popular in the summer. Despite this principle, we note that recent studies on a popular benchmark dataset from Porto, Portugal have found, at best, only marginal improvements in predictive performance from incorporating temporal information. We propose an approach based on hypernetworks, a variant of meta-learning (``learning to learn'') in which a neural network learns to change its own weights in response to an input. In our case, the weights responsible for destination prediction vary with the metadata, in particular the time, of the input trajectory.  The time-conditioned weights notably improve the model's error relative to ablation studies and comparable prior work, and we confirm our hypothesis that knowledge of time should improve prediction of a vehicle's intended destination.
\end{abstract}



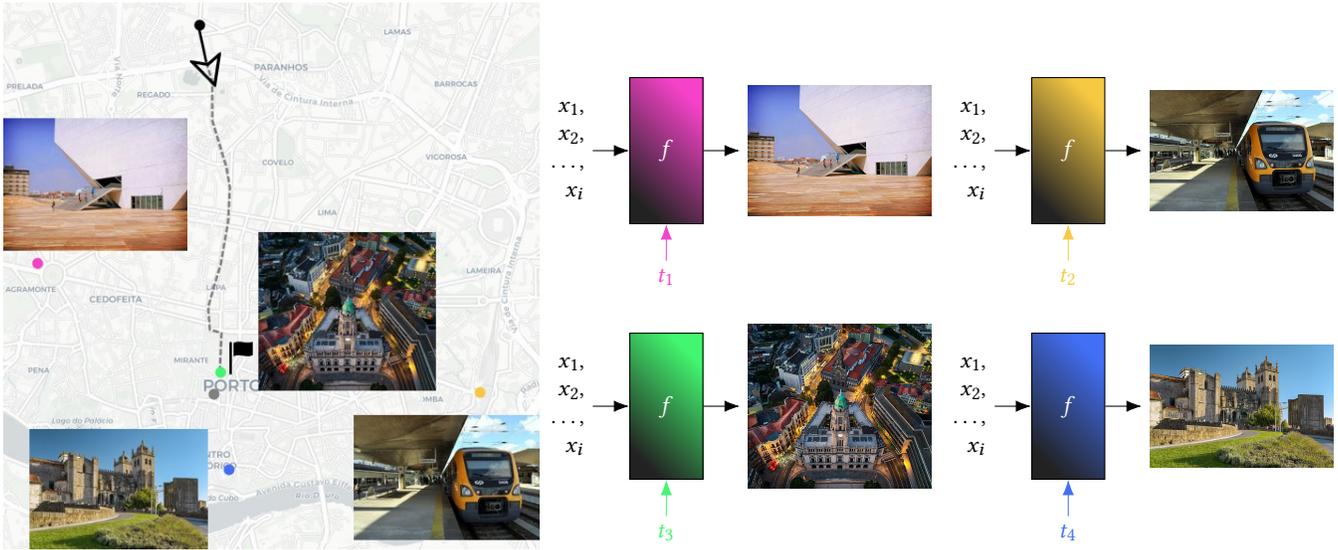
\begin{teaserfigure}
  \includegraphics[width=.4\textwidth]{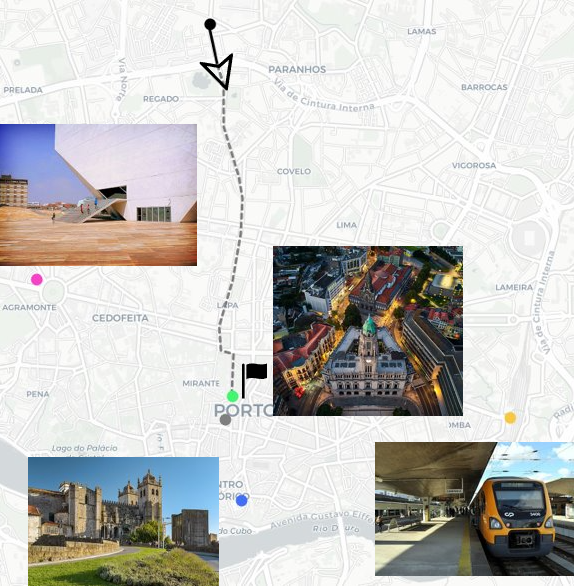}
  \adjustbox{width=.6\textwidth}{\begin{tikzpicture}

    \node[anchor=center, align=center, shading = axis, rectangle,
    left color=left, right color=t1,shading angle=135,
    minimum height =2cm, minimum width = 1cm,
    draw=black, fill=white] (net1) at (1,0) {\textcolor{white}{$f$}};
    
    \node[anchor=right,align=right,left=.5cm of net1] (input1) {$x_1,$\\$x_2,$\\$\ldots,$\\$x_i$};
    \node[anchor=top, below=.5cm of net1] (time1) {\textcolor{t1}{$t_1$}};
    \node[anchor=left, right=.5cm of net1] (output1) {\includegraphics[width=2.5cm]{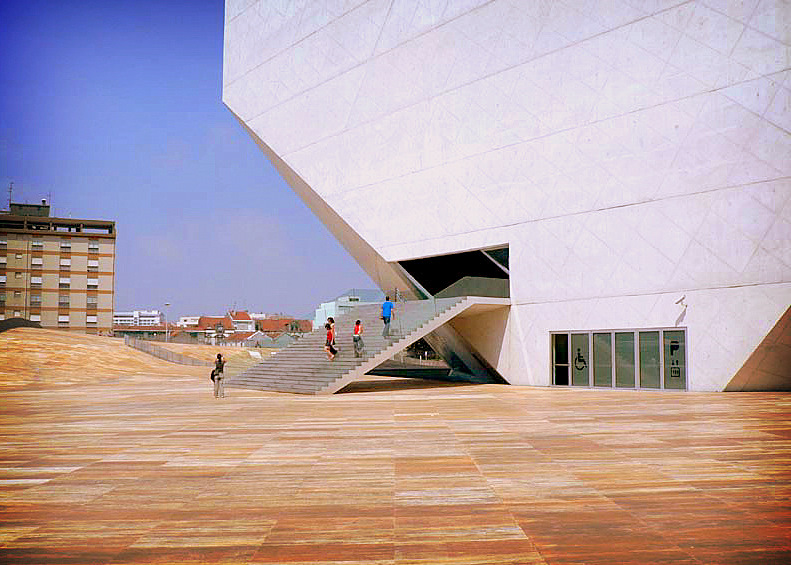}};
    
    \draw[->] (input1) -- (net1);
    \draw[->, , draw=t1] (time1) -- (net1); %
    \draw[->] (net1) -- (output1);
    
    \node[anchor=center, align=center, shading = axis, rectangle,
    left color=left, right color=t2,shading angle=135,
    minimum height =2cm, minimum width = 1cm,
    draw=black, fill=white] (net2) at (6.5,0) {\textcolor{white}{$f$}};
    
    \node[anchor=right,align=right,left=.5cm of net2] (input2) {$x_1,$\\$x_2,$\\$\ldots,$\\$x_i$};
    \node[anchor=top, below=.5cm of net2] (time2) {\textcolor{t2}{$t_2$}};
    \node[anchor=left, right=.5cm of net2] (output2) {\includegraphics[width=2.5cm]{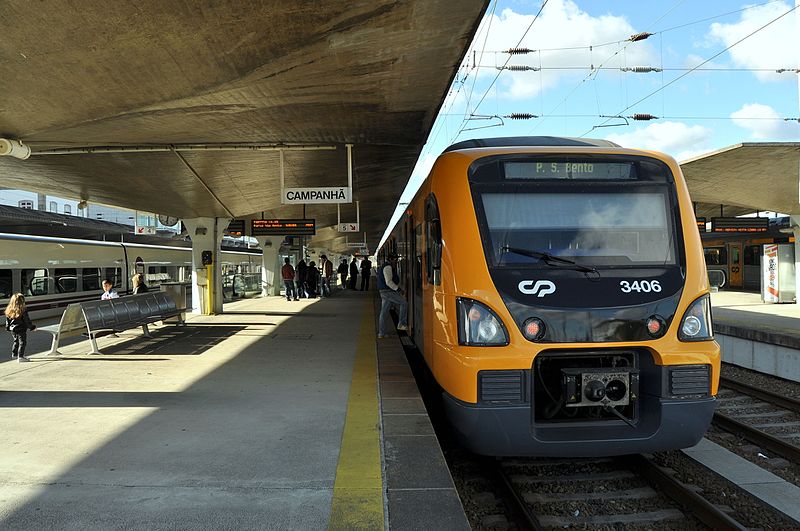}};
    
    \draw[->] (input2) -- (net2);
    \draw[->, , draw=t2] (time2) -- (net2);
    \draw[->] (net2) -- (output2);
    
    \node[anchor=center, align=center, shading = axis, rectangle,
    left color=left, right color=t3,shading angle=135,
    minimum height =2cm, minimum width = 1cm,
    draw=black, fill=white] (net3) at (1, -3.5) {\textcolor{white}{$f$}};
    
    \node[anchor=right,align=right,left=.5cm of net3] (input3) {$x_1,$\\$x_2,$\\$\ldots,$\\$x_i$};
    \node[anchor=top, below=.5cm of net3] (time3) {\textcolor{t3}{$t_3$}};
    \node[anchor=left, right=.5cm of net3] (output3) {\includegraphics[width=2.5cm]{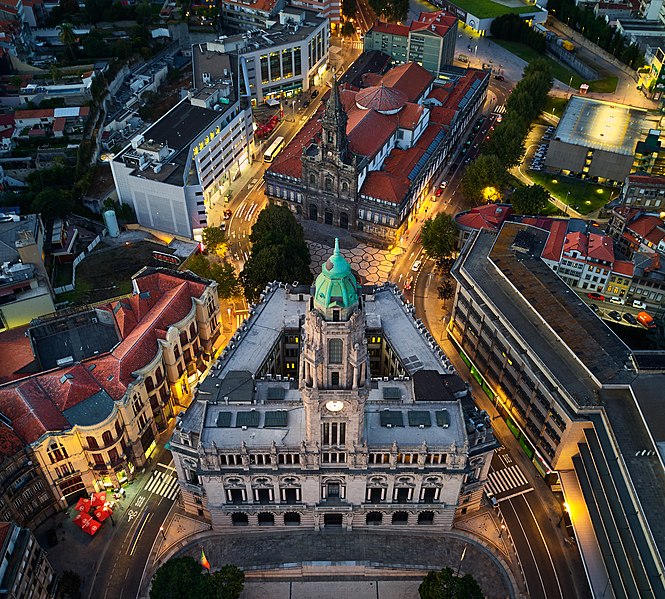}};
    
    \draw[->] (input3) -- (net3);
    \draw[->, , draw=t3] (time3) -- (net3);
    \draw[->] (net3) -- (output3);
    
    \node[anchor=center, align=center, shading = axis, rectangle,
    left color=left, right color=t4,shading angle=135,
    minimum height =2cm, minimum width = 1cm,
    draw=black, fill=white] (net4) at (6.5, -3.5) {\textcolor{white}{$f$}};
    
    \node[anchor=right,align=right,left=.5cm of net4] (input4) {$x_1,$\\$x_2,$\\$\ldots,$\\$x_i$};
    \node[anchor=top, below=.5cm of net4] (time4) {\textcolor{t4}{$t_4$}};
    \node[anchor=left, right=.5cm of net4] (output4) {\includegraphics[width=2.5cm]{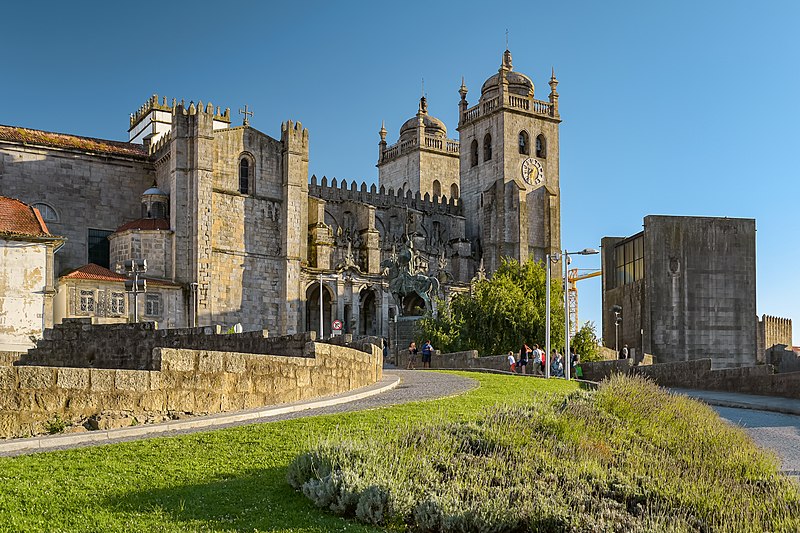}};
    
    \draw[->] (input4) -- (net4);
    \draw[->, , draw=t4] (time4) -- (net4);
    \draw[->] (net4) -- (output4);
    
\end{tikzpicture}
}
  \caption{Where is the taxi passenger going? Based on only a few GPS points, the vehicle could be going to any number of locations, such as the Casa da M{\'u}sica perfomance center (west), the S{\'e} do Porto cathedral (south), or the Estaci{\'o}n de Porto-Campanh{\~a} train station (east), among others. In this case, the model's prediction (gray) is very near the true destination, near the historic C{\^a}mara Municipal do Porto city hall. The current time might provide a hint about the passenger's intentions, and different times of the day, week, or year might imply different destinations. (Photographs under CC BY 2.0, \cite{photo_casadamusica, photo_sedoporto, photo_cityhall, photo_campanha}.)}
  \Description{Where is the taxi passenger going? Based on only a few GPS points, the vehicle could be going to any number of locations, such as the Casa da M{\'u}sica perfomance center (west), the S{\'e} do Porto cathedral (south), or the Estaci{\'o}n de Porto-Campanh{\~a} train station (east), among others. In this case, the model's prediction (gray) is very near the true destination, near the historic C{\^a}mara Municipal do Porto city hall. The current time might provide a hint about the passenger's intentions, and different times of the day, week, or year might imply different destinations.}
  \label{fig:teaser}
\end{teaserfigure}

\maketitle

\section{Introduction}

In recent years, the intersection of transportation and geospatial machine learning has expanded rapidly, fueled in part by a growing number of publicly available GPS datasets. Many are drawn from public transportation systems, especially taxicabs, which provide rich information about mobility patterns and behavior throughout a city while avoiding the privacy pitfall of publishing private vehicles' trajectories.  Common examples include taxis from San Francisco \cite{epfl-mobility-20090224}, New York City \cite{nyctaxi}, Rome \cite{roma-taxi-20140717}, and Beijing \cite{yuan2010t-drive}. Perhaps the most well-known of these benchmarks is the Taxi Service Trajectory prediction challenge from ECML/PKDD 2015 \cite{portowebsite}, which tracked 442 taxis in Porto, Portugal for one year.  Competitors were challenged to predict a taxi's final GPS destination given only a limited ``prefix'' of the first $n$ GPS points and limited metadata about the taxi ride (notably \emph{not} including the total length). More than 380 teams participated in the competition, drawing research interest that continued even after its conclusion \cite{lv2018t, zhang2018multi, rossi2019modelling, zhang2019prediction, li2020fast, ebel2020destination, liao2021taxi, tang2021trip, tsiligkaridis2020personalized}. The Porto dataset has also contributed to other geospatial modeling interests, including passenger demand modeling \cite{moreira2013predicting, saadallah2018bright, le2019neighborhood, rodrigues2020spatiotemporal}, travel time prediction \cite{hoch2015ensemble, gupta2018taxi, fu2019deepist, lan2019travel, das2019map, abbar2020stad}, and unsupervised geospatial learning and anomaly detection \cite{lam2016concise, keane2017detecting, irvine2018normalcy, song2018anomalous, liu2020online}.

Given the successes modeling the primary GPS data, one might expect that the additional trajectory-level features, such as time, might also contain useful knowledge. For example, when modeling destinations, one might expect that Porto's beaches would be busier in summer than winter; recreational areas, more popular on weekends; and restaurants and bars, more popular at mealtimes and evenings.  Moreover, major holidays and events should exhibit learnable deviations from the typical patterns of everyday life. Instead, it is surprising how \emph{unuseful} these metadata are, based on prior researchers' ablation studies.  For example, \citet{ebel2020destination} found that these ancillary features reduced their mean error by merely 0.030 km, or roughly 2 percent of the GPS-only model's error. In a second dataset from San Francisco, incorporating these metadata \emph{increased} the average error between predicted and real destinations \cite{ebel2020destination}.

In this work, we present a more powerful temporal prediction framework drawing on recent work in the meta-learning community. We show, by a series of careful experiments and ablations on the Porto dataset, that: \begin{itemize}
    \item Temporal cues do, in fact, improve performance on taxi destination prediction relative to a model without them, especially when few GPS points are available;
    \item A hypernetwork-based meta-learning framework outperforms the simple concatenatation of temporal and geospatial features commonly used in prior work;
    \item Different timescales (relative to the 24-hour, 7-day, and 1-year cycles) all contribute to this performance, and outperform a time-unaware baseline;
    \item Performance slightly, but noticeably, improves with later rather than earlier fusion. In other words, combining temporal and geospatial encodings works best in the model's final layers, rather than its first ones.
\end{itemize}

\section{Related Work}

The task of destination and/or trajectory prediction is well-known in the domains of urban planning and management, intelligent transportation systems, ``smart cities,'' and ride-sharing applications, leading to a proliferation of research in the past few years \cite{altshuler2019modeling, anagnostopoulos2021predictive, zhao2019data, zheng2021rebuilding, xiao2020extracting, wang2020attention}. Recent events have also energized a sub-field of mobility modeling and destination prediction for the purposes of pandemic analysis \cite{jiang2021countrywide}.  Many of these models fall broadly under the neural network framework, although they vary widely in their complexity. In the context of Porto, early successes were achieved by relatively simple multi-layer perceptron (MLP) models, such as the competition-winning result of \citet{de2015artificial}, which outperformed competitors such as ensembles of regression trees \cite{lam2015blue}. More recent papers have predominantly employed recurrent networks instead \cite{rossi2019modelling, liao2021taxi, li2020fast, ebel2020destination, zhang2018multi, tang2021trip}, particularly long short-term memory (LSTM) networks \cite{hochreiter1997long}. Notable exceptions include \citet{lv2018t}, who plot trajectories graphically and then model them as 2D images rather than GPS points or embeddings, using convolutional networks, and \citet{tsiligkaridis2020personalized}, who use Transformers \cite{vaswani2017attention}.

Within the LSTM-based community, there remains significant diversity in the model architectures. \citet{zhang2018multi} combine image-based CNN and GPS recurrent representations into a single predictive model. \citet{ebel2020destination} and \citet{liao2021taxi} both perform a geospatial partitioning of the Porto area and train a LSTM over a sequence of region embeddings; \citet{ebel2020destination} use $k$-d trees, while \citet{liao2021taxi} use grid-based and quadtree partitions and add an attention mechanism. In a unique approach, \citet{rossi2019modelling} consider a \emph{driver}'s history rather than a passenger's, modeling a sequence of pick-up/drop-off points across consecutive trajectories, rather than each trajectory individually. Our own approach is most closely related to \citet{ebel2020destination} and \citet{liao2021taxi}, in that we learn geospatial embeddings over GPS points and model trajectories as sequences of embeddings.  However, the details of our approach, including our embedding mechanism, temporal representation, and model architecture, differ signficantly (see \cref{sec:geospatial_encoding,sec:metadata,sec:hypernet_comparison}).

\section{Methods}

\subsection{Problem Definition}

\subsubsection{Trajectory.} Let $\mathcal{T}$ be the set of all trajectories, where each trajectory $T:=((p_i)^{N}_{i=1}, M)$ is defined as a sequence of points $p_1, \ldots, p_N$ where $N$ is the length of the trajectory and metadata $M$ associated with the trajectory. Each point $p_i = (\phi_i, \lambda_i)$ is an ordered pair of latitude and longitude values, and the destination of a trajectory is its final point $p_N$.

\subsubsection{Trajectory prefix.} A partial trajectory, or trajectory fragment, is some contiguous sub-sequence of points $(p_i)^b_a = p_a, \ldots, p_b$ for some lower and upper indices $a,b$. We will refer to the $n$-length \emph{prefix} of trajectory $T^{(n)}$ as the partial trajectory which starts at the beginning of the trajectory $p_1$ and is $n$ points long, where by assumption $1 \leq n \leq N$.

\subsubsection{Metadata.} The metadata $M$ are broadly defined as any additional information associated with a trajectory, such as an identifier for each taxi driver. Of particular interest is the timestamp $t$ (taken, in the Porto dataset, from the beginning of the trajectory). We hypothesize that the behavioral patterns of both drivers and passengers should vary with the time of day, day of the week, and season.

\subsubsection{Haversine distance.} The Haversine distance measures the great-circle distance between two points on a sphere, i.e., the shortest distance between two points traveling along the sphere's surface.  Following the original competition metric, the distance between two points is
\begin{align}
    d_H(p_i, p_j) &= 2 r \arctan \left( \sqrt{\frac{a}{1-a}} \right) \textrm{, with} \\
    a &= \sin^2 \left( \frac{\phi_j - \phi_i}{2} \right) + \cos \phi_i \cos \phi_j \sin^2 \left( \frac{\lambda_j - \lambda_i}{2} \right)
\end{align}

\begin{figure*}
  \centering
  \adjustbox{width=\textwidth}{\input{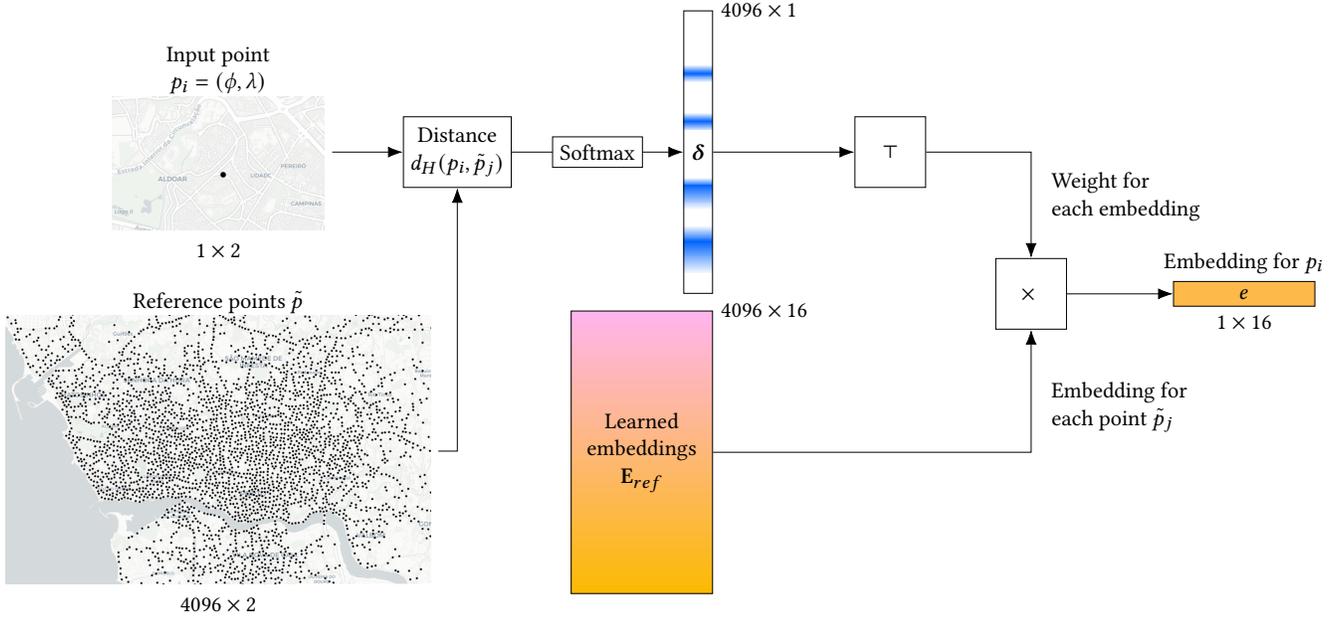}}
  \caption{Input encoding mechanism, converting GPS points to an embedding space. In contrast to prior work, we avoid performing a ``hard assignment'' or discrete partition of the geospatial area or road network. Instead, we learn embeddings for each of a set of 4,096 reference points, and each input point is encoded as a ``soft'' distance-weighted average of these embeddings.}
  \label{fig:geospatial_encoding}
\end{figure*}

\subsubsection{Destination prediction} Our problem, then, is as follows: for any $n$, given a trajectory prefix $T^{(n)} = (p_1, \ldots, p_n)$ and metadata $M$, and predict the final destination $p_N$, such that the Haversine distance between the predicted and real $p_N$ is minimized. Notably, the prefix length $n$ is known but the length of the \emph{full} trajectory $N$ is not, except that by definition $N \geq n$. In other words, the model must infer the final destination without being informed how much time or distance is remaining.

\subsection{Data Preprocessing}

For the purposes of fair comparison, we follow the preprocessing procedure outlined by \citet{ebel2020destination}:

\begin{enumerate}
    \item We remove trajectories that are extremely short (2 minutes or less) or long (2 hours or longer), including those consisting of only a single data point.
    \item If the apparent speed between two points is extremely large, we assume GPS measurement error. Whenever the speed greater than 240 km/h, we smooth the outliers with a median filter.
    \item Any remaining trajectories which include points outside the Porto area are removed.
    \item We remove roundtrips or sight-seeing trips, i.e., a long trajectory which starts and ends near the same point, traversing through points of interest, especially tourist ones. These trips only confound the problem of destination prediction. The roundtrip factor $\tau$ \cite{ebel2020destination} is the ratio of the trip's total path length to the bee-line distance between its start and end.  If $\tau$ is large, then the trajectory travels a long path but ends close to where it began.
    \begin{equation}
        \tau = \frac{\sum_i d_H(p_{i}, p_{i+1})}{d_H(p_1, p_N)}
    \end{equation}
    Following \citet{ebel2020destination}, we only keep trajectories with $\tau < 3.5$ (roughly the 95th percentile of $\tau$), and drop the remaining ones from the dataset.
\end{enumerate}

\subsection{Geospatial Encoding}
\label{sec:geospatial_encoding}

\begin{figure*}[H]
  \centering
  \adjustbox{width=\textwidth}{\begin{tikzpicture}
\node[anchor=center, align=center, minimum width=2cm] (point1) at (0,0) {
    $p_1 = \phi_1, \lambda_1$
};
\node[anchor=center, align=center, below=.5cm of point1, minimum width=2cm] (point2) {
    $p_2 = \phi_2, \lambda_2$
};
\node[anchor=center, align=center, below=.5cm of point2, minimum width=2cm] (point3) {
    $\vdots$
};
\node[anchor=center, align=center, below=.5cm of point3, minimum width=2cm] (point4) {
    $p_N = \phi_N, \lambda_N$
};

\node[draw, dotted,inner sep=2mm, fit=(point1) (point2) (point3) (point4)] (trajectory) {};
\node[anchor=center, align=center, below=0cm of trajectory] {Input\\trajectory\\$N \times 2$};

\node[anchor=center, align=center, right=2.cm of point1, draw=black, fill=pink, minimum width=2cm] (embed1) {$e_1$};
\draw[->] (point1) -- (embed1);
\node[anchor=center, align=center, right=2.cm of point2, minimum size=.5cm, draw=black, fill=pink!60!orange, minimum width=2cm] (embed2) {$e_2$};
\draw[->] (point2) -- (embed2);
\node[anchor=center, align=center, minimum width=2cm] (embed3) at ($(point3 -| embed2)$) {$\vdots$};
\draw[->] (point3) -- (embed3);
\node[anchor=center, align=center, right=2cm of point4, draw=black, fill=orange, minimum width=2cm] (embed4) {$e_N$};
\draw[->] (point4) -- (embed4);
\node[draw, dotted,inner sep=2mm,fit=(embed1) (embed2) (embed3) (embed4)] (embeddings) {};
\node[anchor=center, align=center, below=0cm of embeddings] {Geospatial\\encoding\\(\cref{sec:geospatial_encoding,fig:geospatial_encoding})\\$N \times 16$};

\node[anchor=center, align=center, right=2.cm of embed1, draw=black, fill=t3] (rnn1) {$f$};
\node[anchor=center, align=center, right=2.cm of embed2, draw=black, fill=t3] (rnn2) {$f$};
\node[anchor=center, align=center] at ($(embed3 -| rnn2)$) (rnn3) {$\vdots$};
\node[anchor=center, align=center, right=2.cm of embed4, draw=black, fill=t3] (rnn4) {$f$};
\draw[->] (embed1) -- (rnn1);
\draw[->] (embed2) -- (rnn2);
\draw[->] (embed3) -- (rnn3);
\draw[->] (embed4) -- (rnn4);
\draw[->] (rnn1) -- (rnn2);
\draw[->] (rnn2) -- (rnn3);
\draw[->] (rnn3) -- (rnn4);

\node[draw, dotted,inner sep=2mm,fit=(rnn1) (rnn2) (rnn3) (rnn4)] (rnn) {};
\node[anchor=center, align=center, below=0cm of rnn] {RNN layers $f$\\(\cref{sec:hypernetworks,sec:hypernet_comparison},\\\cref{fig:hypermodels})};

\node[anchor=center, align=center, above=0.5cm of rnn1] (meta) {Metadata $M$\\(\cref{sec:metadata})};
\draw[->] (meta) -- (rnn1);

\node[anchor=center, align=center, right=2.cm of rnn1, draw=black, shading=delta_shade, minimum width=2cm] (alpha1) {$\boldsymbol{\alpha}_1$};
\draw[->] (rnn1) -- (alpha1);
\node[anchor=center, align=center, right=2.cm of rnn2, draw=black, shading=delta_shade, minimum width=2cm] (alpha2) {$\boldsymbol{\alpha}_2$};
\draw[->] (rnn2) -- (alpha2);
\node[anchor=center, align=center, minimum width=2cm] (alpha3) at ($(rnn3 -| alpha2)$) {$\vdots$};
\draw[->] (rnn3) -- (alpha3);
\node[anchor=center, align=center, right=2.cm of rnn4, draw=black, shading=delta_shade, minimum width=2cm] (alpha4) {$\boldsymbol{\alpha}_N$};
\draw[->] (rnn4) -- (alpha4);
\node[draw, dotted,inner sep=2mm,fit=(alpha1) (alpha2) (alpha3) (alpha4)] (alpha) {};
\node[anchor=center, align=center, below=0cm of alpha] {Softmax\\weights\\(\cref{sec:output})\\$N \times 4096$};

\node[anchor=center, align=center, right=2.cm of alpha1,  minimum width=2cm] (yhat1) {$\hat{y}_1 = (\hat{\phi}_1, \hat{\lambda}_1)$};
\draw[->] (alpha1) -- (yhat1);
\node[anchor=center, align=center, right=2.cm of alpha2,  minimum width=2cm] (yhat2) {$\hat{y}_2 = (\hat{\phi}_2, \hat{\lambda}_2)$};
\draw[->] (alpha2) -- (yhat2);
\node[anchor=center, align=center, right=2.cm of alpha3,  minimum width=2cm] (yhat3) {$\vdots$};
\draw[->] (alpha3) -- (yhat3);
\node[anchor=center, align=center, right=2.cm of alpha4,  minimum width=2cm] (yhat4) {$\hat{y}_N = (\hat{\phi}_N, \hat{\lambda}_N)$};
\draw[->] (alpha4) -- (yhat4);
\node[draw, dotted,inner sep=2mm,fit=(yhat1) (yhat2) (yhat3) (yhat4)] (yhat) {};
\node[anchor=center, align=center, below=0cm of yhat] {Predicted destination\\for each prefix\\$N \times 2$\\$\forall i$, $ \hat{y}_i \rightarrow p_N$};

\node[anchor=center, align=center, right=2.cm of yhat4, draw=red] (pn) {\textcolor{red}{$p_N$}};
\draw[<-, color=red] (yhat1.east) -| (pn) node[midway, right] {$\mathcal{L}$};
\draw[<-, color=red] (yhat2.east) -| (pn) node[midway, right] {$\mathcal{L}$};
\draw[<-, color=red] (yhat3.east) -| (pn) node[midway, right] {$\mathcal{L}$};
\draw[<-, color=red] (yhat4.east) -- (pn) node[midway, below] {$\mathcal{L}$};

\end{tikzpicture}}
  \caption{Overall information flow of the model. A sequence of points is converted to a sequence of embeddings via a geospatial encoding mechanism (\cref{sec:geospatial_encoding}). This sequence is modeled by fully-connected and recurrent layers (\cref{sec:hypernet_comparison}), some parameterized by a hypernetwork (\cref{sec:hypernetworks}). The output is a softmax weight vector over the reference points from the geospatial encoding, which is simple to convert to a predicted point (\cref{sec:output}). Thus, for every $i = 1, \ldots, N$ we have a prediction $\hat{y}_i$ based on points $p_1, \ldots, p_i$ trained to match the final destination $p_N$.}
  \label{fig:overall_model}
\end{figure*}

Most destination prediction models begin by partitioning the geospatial area into a set of regions. The ground-truth road network itself may be partitioned into segments \cite{neto2018combining, lassoued2017hidden, li2016t, simmons2006learning}. Alternatively, the geospatial area can be divided into uniform grids \cite{krumm2006predestination, manasseh2013predicting, endo2017predicting, pecher2016data} or, with greater sophistication, into $k$-d tree-based regions so that smaller, more precise regions are used in areas with higher vehicle density \cite{xue2015solving, ebel2020destination}. (See \cite{ebel2020destination} for related discussion.) We instead propose an approach which does not explicitly partition the input GPS space at all (\cref{fig:geospatial_encoding}).

We randomly sample GPS points from training trajectories. At each random draw, we find the nearest neighbor in our current sample, and if this minimum distance is at least $d_H \geq 0.1$ km we add the point to our sample. We repeat until we have sampled $N_{ref} = 4096$ points, which we refer to as the \emph{reference points}, or $\tilde{p}$.

We instantiate an embedding table $\mathbf{E}_{ref}$ of shape $N_{ref} \times m$, that is, a table of $N_{ref}$ different embedding vectors of size $m$. These embeddings $\mathbf{E}_{ref}$ are learnable model parameters trained jointly with the rest of the model. To convert the points of a trajectory $T$ to embeddings, we consider each point $p_i = (\phi_i, \lambda_i)$ separately.  We calculate the Haversine distance $d_H$ from $p_i$ to each reference point $\tilde{p}_j$.  We then negate these distances and apply a softmax function, creating a $N_{ref} \times 1$ vector $\boldsymbol \delta$
\begin{equation}
    \delta_j = \frac{\exp(-d_H(p_i, \tilde{p}_j))}{\sum_k \exp(-d_H(p_i, \tilde{p}_k))}
\end{equation}
so that $\delta_j$ is largest for reference points $\tilde{p}_j$ closest to the point $p_i$, and low for faraway points. Due to the softmax, $\boldsymbol \delta$ always sums to 1.

The embedding of point $p_i$ is an average of the embeddings of all the reference points, weighted by the proximity $\boldsymbol \delta$: 
\begin{equation}
    e = \boldsymbol \delta^{\top} \mathbf{E}_{ref}
\end{equation}
This converts the $N \times 2$ sequence of points into a $N \times m$ sequence of embeddings.  We find that a relatively small embedding of size $m = 16$ works well. %

\subsection{Temporal Encoding}
\label{sec:metadata}

We represent time as a set of continuous oscillating functions: the time of day, week, and year.  Each takes the form of a sinusoidal embedding, which converts a timestamp (in hours) $t$ to a vector:
\begin{align}
    c(t) &= \sin \left( \frac{2 \pi}{C} t + \boldsymbol 
    \varphi \right) \mathrm{, with} \\
    \boldsymbol \varphi &= \left[ 0, \frac{\pi}{2}, \pi, \frac{3\pi}{2} \right]
\end{align}
where $C$ is the period in hours, i.e., 24, 168, and 8760 for a day, week, and year respectively.

\subsubsection{Other metadata.} The Porto dataset also includes additional categorical metadata: (i) a unique ID anonymously identifying the customer's phone number, if available, (ii) a unique ID identifying the taxi stand where the trip began, if applicable, and (iii) a unique ID for the taxi driver. We incorporate these metadata features as learned embeddings, as has long been popular for this dataset \cite{de2015artificial}, and concatenate them with the three temporal encodings $c(t)$.

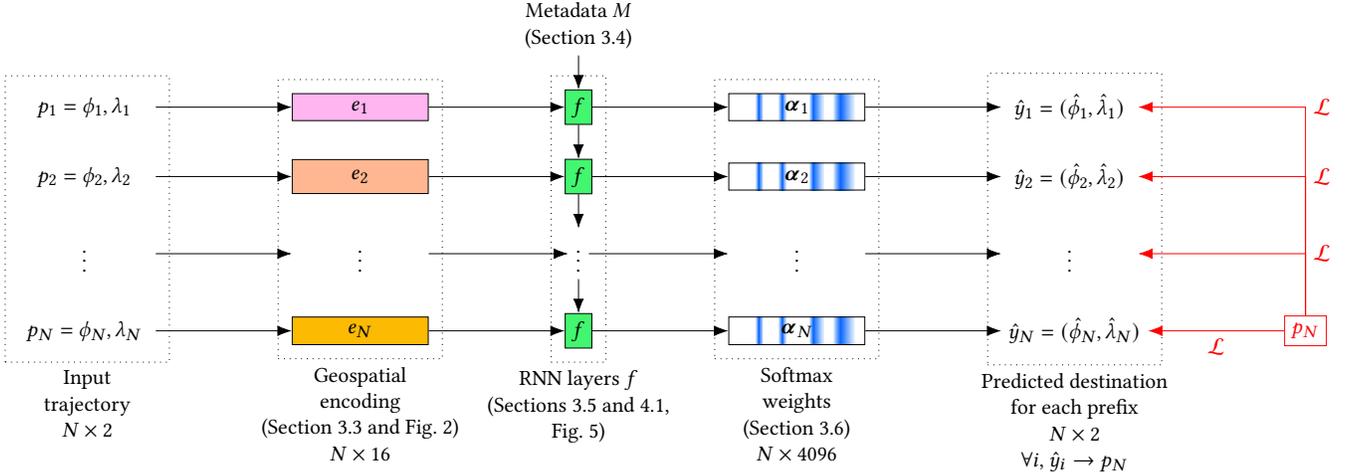
\begin{figure*}
  \centering
  \adjustbox{width=\textwidth}{\begin{tikzpicture}
\node[anchor=center, align=center, minimum width=2cm] (point1) at (0,0) {
    $p_1 = \phi_1, \lambda_1$
};
\node[anchor=center, align=center, below=.5cm of point1, minimum width=2cm] (point2) {
    $p_2 = \phi_2, \lambda_2$
};
\node[anchor=center, align=center, below=.5cm of point2, minimum width=2cm] (point3) {
    $\vdots$
};
\node[anchor=center, align=center, below=.5cm of point3, minimum width=2cm] (point4) {
    $p_N = \phi_N, \lambda_N$
};

\node[draw, dotted,inner sep=2mm, fit=(point1) (point2) (point3) (point4)] (trajectory) {};
\node[anchor=center, align=center, below=0cm of trajectory] {Input\\trajectory\\$N \times 2$};

\node[anchor=center, align=center, right=2.cm of point1, draw=black, fill=pink, minimum width=2cm] (embed1) {$e_1$};
\draw[->] (point1) -- (embed1);
\node[anchor=center, align=center, right=2.cm of point2, minimum size=.5cm, draw=black, fill=pink!60!orange, minimum width=2cm] (embed2) {$e_2$};
\draw[->] (point2) -- (embed2);
\node[anchor=center, align=center, minimum width=2cm] (embed3) at ($(point3 -| embed2)$) {$\vdots$};
\draw[->] (point3) -- (embed3);
\node[anchor=center, align=center, right=2cm of point4, draw=black, fill=orange, minimum width=2cm] (embed4) {$e_N$};
\draw[->] (point4) -- (embed4);
\node[draw, dotted,inner sep=2mm,fit=(embed1) (embed2) (embed3) (embed4)] (embeddings) {};
\node[anchor=center, align=center, below=0cm of embeddings] {Geospatial\\encoding\\(\cref{sec:geospatial_encoding,fig:geospatial_encoding})\\$N \times 16$};

\node[anchor=center, align=center, right=2.cm of embed1, draw=black, fill=t3] (rnn1) {$f$};
\node[anchor=center, align=center, right=2.cm of embed2, draw=black, fill=t3] (rnn2) {$f$};
\node[anchor=center, align=center] at ($(embed3 -| rnn2)$) (rnn3) {$\vdots$};
\node[anchor=center, align=center, right=2.cm of embed4, draw=black, fill=t3] (rnn4) {$f$};
\draw[->] (embed1) -- (rnn1);
\draw[->] (embed2) -- (rnn2);
\draw[->] (embed3) -- (rnn3);
\draw[->] (embed4) -- (rnn4);
\draw[->] (rnn1) -- (rnn2);
\draw[->] (rnn2) -- (rnn3);
\draw[->] (rnn3) -- (rnn4);

\node[draw, dotted,inner sep=2mm,fit=(rnn1) (rnn2) (rnn3) (rnn4)] (rnn) {};
\node[anchor=center, align=center, below=0cm of rnn] {RNN layers $f$\\(\cref{sec:hypernetworks,sec:hypernet_comparison},\\\cref{fig:hypermodels})};

\node[anchor=center, align=center, above=0.5cm of rnn1] (meta) {Metadata $M$\\(\cref{sec:metadata})};
\draw[->] (meta) -- (rnn1);

\node[anchor=center, align=center, right=2.cm of rnn1, draw=black, shading=delta_shade, minimum width=2cm] (alpha1) {$\boldsymbol{\alpha}_1$};
\draw[->] (rnn1) -- (alpha1);
\node[anchor=center, align=center, right=2.cm of rnn2, draw=black, shading=delta_shade, minimum width=2cm] (alpha2) {$\boldsymbol{\alpha}_2$};
\draw[->] (rnn2) -- (alpha2);
\node[anchor=center, align=center, minimum width=2cm] (alpha3) at ($(rnn3 -| alpha2)$) {$\vdots$};
\draw[->] (rnn3) -- (alpha3);
\node[anchor=center, align=center, right=2.cm of rnn4, draw=black, shading=delta_shade, minimum width=2cm] (alpha4) {$\boldsymbol{\alpha}_N$};
\draw[->] (rnn4) -- (alpha4);
\node[draw, dotted,inner sep=2mm,fit=(alpha1) (alpha2) (alpha3) (alpha4)] (alpha) {};
\node[anchor=center, align=center, below=0cm of alpha] {Softmax\\weights\\(\cref{sec:output})\\$N \times 4096$};

\node[anchor=center, align=center, right=2.cm of alpha1,  minimum width=2cm] (yhat1) {$\hat{y}_1 = (\hat{\phi}_1, \hat{\lambda}_1)$};
\draw[->] (alpha1) -- (yhat1);
\node[anchor=center, align=center, right=2.cm of alpha2,  minimum width=2cm] (yhat2) {$\hat{y}_2 = (\hat{\phi}_2, \hat{\lambda}_2)$};
\draw[->] (alpha2) -- (yhat2);
\node[anchor=center, align=center, right=2.cm of alpha3,  minimum width=2cm] (yhat3) {$\vdots$};
\draw[->] (alpha3) -- (yhat3);
\node[anchor=center, align=center, right=2.cm of alpha4,  minimum width=2cm] (yhat4) {$\hat{y}_N = (\hat{\phi}_N, \hat{\lambda}_N)$};
\draw[->] (alpha4) -- (yhat4);
\node[draw, dotted,inner sep=2mm,fit=(yhat1) (yhat2) (yhat3) (yhat4)] (yhat) {};
\node[anchor=center, align=center, below=0cm of yhat] {Predicted destination\\for each prefix\\$N \times 2$\\$\forall i$, $ \hat{y}_i \rightarrow p_N$};

\node[anchor=center, align=center, right=2.cm of yhat4, draw=red] (pn) {\textcolor{red}{$p_N$}};
\draw[<-, color=red] (yhat1.east) -| (pn) node[midway, right] {$\mathcal{L}$};
\draw[<-, color=red] (yhat2.east) -| (pn) node[midway, right] {$\mathcal{L}$};
\draw[<-, color=red] (yhat3.east) -| (pn) node[midway, right] {$\mathcal{L}$};
\draw[<-, color=red] (yhat4.east) -- (pn) node[midway, below] {$\mathcal{L}$};

\end{tikzpicture}}
  \caption{Overall information flow of the model. A sequence of points is converted to a sequence of embeddings via a geospatial encoding mechanism (\cref{sec:geospatial_encoding}). This sequence is modeled by fully-connected and recurrent layers (\cref{sec:hypernet_comparison}), some parameterized by a hypernetwork (\cref{sec:hypernetworks}). The output is a softmax weight vector over the reference points from the geospatial encoding, which is simple to convert to a predicted point (\cref{sec:output}). Thus, for every $i = 1, \ldots, N$ we have a prediction $\hat{y}_i$ based on points $p_1, \ldots, p_i$ trained to match the final destination $p_N$.}
  \label{fig:overall_model}
\end{figure*}

\subsection{Hypernetwork and LSTM}
\label{sec:hypernetworks}

Given a neural network $f_\theta(x) \rightarrow y$, that is, a network with weights $\theta$ trained to predict $y$ from $x$, a hypernetwork \cite{ha2016hypernetworks} refers to a network
\begin{equation}
    h_\gamma: Z \rightarrow \theta
\end{equation}
which learns to generate the main network's weights $\theta$ from some input $Z$. Because all operations in both $f$ and $h$ are differentiable, all learnable parameters can be trained via backpropagation. In practice, generating all parameters $\theta$ for an entire neural network is prohibitively expensive, and instead a subset of $\theta$ are generated by $h$. Recent authors have argued that this framing allows for a more powerful information fusion mechanism for two input streams in a neural network \cite{jayakumar2019multiplicative, galanti2020modularity} and suggest it as a drop-in replacement for concatenation.  Outside the geospatial community, hypernetworks and related methods have found success in diverse applications including neural style transfer \cite{shen2018neural}, time series analysis \cite{deng2020variational}, multi-task \cite{mahabadi2021parameter} and continual learning \cite{von2019continual}, robotic driving applications \cite{wang2018real}, reinforcement learning \cite{sarafian2021recomposing}, and medical prediction \cite{ji2021patient}.  In our setting, we consider the case where the hypernetwork $h$ is a single fully-connected (linear) layer operating over the inputs described in \cref{sec:metadata}, that is, $Z$ consists of the temporal encodings $c(t)$, the driver and customer ID, and the taxi stand if applicable. The architecture of the main network $f_\theta$ varies in our experiments (see \cref{sec:hypernet_comparison}) but may be a recurrent network, such as LSTM \cite{hochreiter1997long}, or another fully-connected layer.

The fully-connected layer, $\mathbf{W} \mathbf{x} + \mathbf{b}$, is the simpler case. Here, we simply have $\theta := \{ \mathbf{W}, \mathbf{b} \}$, that is, we generate a weight matrix and bias vector from a hypernetwork $h_\gamma: Z \rightarrow \theta$. In our formulation, $\mathbf{W}$ and $\mathbf{b}$ are each generated by a fully-connected layer, so this hypernetwork $h$ is functionally equivalent to the \emph{multiplicative interaction} discussed in detail by \citet{jayakumar2019multiplicative}. The LSTM module \cite{hochreiter1997long} internally can be viewed as four fully-connected layers (with appropriate non-linearities applied), so to paramaterize a LSTM via hypernetwork, we simply expand $\theta$ to generate four weight matrices and biases rather than one.
\subsubsection{Weight Normalization}

However, hypernetwork training can be difficult in practice; the optimization may be unstable, or may not converge at all.  Hypernetworks, especially at initialization, often produce weights with substantially larger or smaller scales than necessary for training, leading to poor results. For specific model architectures, proper weight initialization schemes can be derived analytically \cite{chang2019principled}, but they do not hold in general (e.g., for both linear and recurrent layers).

Similar to \citet{krueger2017bayesian}, we find that applying weight normalization \cite{salimans2016weight} consistently succeeds in stabilizing the hypernetwork training process.  For a weight vector $\mathbf{v} \in \theta$ generated by $h$, the normalized weight $\mathbf{w}$ is simply
\begin{equation}
    \mathbf{w} = \frac{g}{\Vert \mathbf{v} \Vert} \mathbf{v}
\end{equation}
where $g$ is a scalar learned via backpropagation. The generated weight $\mathbf{v}$ is normalized and multiplied by this learned constant, guaranteeing that the scale of $\mathbf{w}$ is always $g$. In other words, the hypernetwork generates the \emph{direction} of this vector, but not its scale. Since generating weights on the wrong scale is the cause of poor training in the first place, this simple reparameterization trick resolves any instability in training and enables us to train hypernetworks $h$ for both fully-connected and LSTM networks $f$ without further interventions, such as custom weight initialization schemes.

\begin{figure*}
  \centering
  \begin{subfigure}{0.3\textwidth}
    \adjustbox{width=\textwidth}{\begin{tikzpicture}
\node[anchor=center, align=center, draw=black, minimum width=2cm] (embed) at (0,0) {$e_i$};

\node[anchor=center, align=center, draw=black, left color=t3, right color=white, shading angle=135, minimum height=2cm, right=1cm of embed] (prelinear) {\rotatebox{90}{Linear}};
\draw[->] (embed) -- (prelinear) node[midway, above] {16};

\node[anchor=center, align=center, draw=black, minimum size=2cm, right=1cm of prelinear] (lstm) {LSTM};
\node[anchor=center, align=center, above=1cm of lstm] (above_lstm) {$\vdots$};
\node[anchor=center, align=center, below=1cm of lstm] (below_lstm) {$\vdots$};
\draw[->] (prelinear) -- (lstm) node[midway, above] {64};
\draw[->] (above_lstm) -- (lstm);
\draw[->] (lstm) -- (below_lstm);

\node[anchor=center, align=center, draw=black, minimum height=2cm, right=1cm of lstm] (final_linear1) {\rotatebox{90}{Linear}};
\node[anchor=center, align=center, draw=black, minimum height=2cm, right=1cm of final_linear1] (final_linear2) {\rotatebox{90}{Linear}};
\node[right=1cm of final_linear2] (alpha) {};

\draw[->] (lstm) -- (final_linear1) node[midway, above] {64};
\draw[->] (final_linear1) -- (final_linear2) node[midway, above] {128};
\draw[->] (final_linear2) -- (alpha) node[midway, above] {4096};

\node[anchor=center, align=center, draw=black, left color=t3, right color=white, shading angle=135, minimum height=2cm, below left=1cm of prelinear.west] (hypernet) {$h$};
\node[anchor=center, align=center, draw=black, left color=t3, right color=white, shading angle=135] (meta) at ($(hypernet -| embed)$) {$M$};

\draw[->] (meta) -- (hypernet);
\draw[->] (hypernet.east) -- (prelinear.south west);

\end{tikzpicture}}
    \caption{Pre-LSTM hypernetwork. \label{fig:hypermodels:a}}
  \end{subfigure}
  \begin{subfigure}{0.3\textwidth}
    \adjustbox{width=\textwidth}{\begin{tikzpicture}
\node[anchor=center, align=center, draw=black, minimum width=2cm] (embed) at (0,0) {$e_i$};

\node[anchor=center, align=center, draw=black, minimum size=2cm, right=1cm of embed, left color=t3, right color=white, shading angle=135] (lstm) {LSTM};
\node[anchor=center, align=center, above=1cm of lstm] (above_lstm) {$\vdots$};
\node[anchor=center, align=center, below=1cm of lstm] (below_lstm) {$\vdots$};
\draw[->] (embed) -- (lstm) node[midway, above] {16};
\draw[->] (above_lstm) -- (lstm);
\draw[->] (lstm) -- (below_lstm);

\node[anchor=center, align=center, draw=black, minimum height=2cm, right=1cm of lstm] (final_linear1) {\rotatebox{90}{Linear}};
\node[anchor=center, align=center, draw=black, minimum height=2cm, right=1cm of final_linear1] (final_linear2) {\rotatebox{90}{Linear}};
\node[right=1cm of final_linear2] (alpha) {};

\draw[->] (lstm) -- (final_linear1) node[midway, above] {64};
\draw[->] (final_linear1) -- (final_linear2) node[midway, above] {128};
\draw[->] (final_linear2) -- (alpha) node[midway, above] {4096};

\node[anchor=center, align=center, draw=black, left color=t3, right color=white, shading angle=135, minimum height=2cm, below left=1cm of lstm.west] (hypernet) {$h$};
\node[anchor=center, align=center, draw=black, left color=t3, right color=white, shading angle=135] (meta) at ($(hypernet -| embed)$) {$M$};

\draw[->] (meta) -- (hypernet);
\draw[->] (hypernet.east) -- (lstm.south west);

\end{tikzpicture}}
    \caption{Hyper-LSTM. \label{fig:hypermodels:b}}
  \end{subfigure}
  \begin{subfigure}{0.3\textwidth}
    \adjustbox{width=\textwidth}{\begin{tikzpicture}
\node[anchor=center, align=center, draw=black, minimum width=2cm] (embed) at (0,0) {$e_i$};

\node[anchor=center, align=center, draw=black, minimum size=2cm, right=1cm of embed] (lstm) {LSTM};
\node[anchor=center, align=center, above=1cm of lstm] (above_lstm) {$\vdots$};
\node[anchor=center, align=center, below=1cm of lstm] (below_lstm) {$\vdots$};
\draw[->] (embed) -- (lstm) node[midway, above] {16};
\draw[->] (above_lstm) -- (lstm);
\draw[->] (lstm) -- (below_lstm);

\node[anchor=center, align=center, draw=black, minimum height=2cm, right=1cm of lstm, left color=t3, right color=white, shading angle=135] (post_linear) {\rotatebox{90}{Linear}};

\node[anchor=center, align=center, draw=black, minimum height=2cm, right=1cm of post_linear] (final_linear1) {\rotatebox{90}{Linear}};
\node[anchor=center, align=center, draw=black, minimum height=2cm, right=1cm of final_linear1] (final_linear2) {\rotatebox{90}{Linear}};
\node[right=1cm of final_linear2] (alpha) {};

\draw[->] (lstm) -- (post_linear) node[midway, above] {64};
\draw[->] (post_linear) -- (final_linear1) node[midway, above] {64};
\draw[->] (final_linear1) -- (final_linear2) node[midway, above] {128};
\draw[->] (final_linear2) -- (alpha) node[midway, above] {4096};

\node[anchor=center, align=center, draw=black, left color=t3, right color=white, shading angle=135, minimum height=2cm, below left=.66cm of post_linear.west] (hypernet) {$h$};
\node[anchor=center, align=center, draw=black, left color=t3, right color=white, shading angle=135, left=.3cm of hypernet] (meta) {$M$};

\draw[->] (meta) -- (hypernet);
\draw[->] (hypernet.east) -- (post_linear.south west);

\end{tikzpicture}}
    \caption{Post-LSTM hypernetwork. \label{fig:hypermodels:c}}
  \end{subfigure}
  \caption{Fusion may occur before the recurrent network, altering the geospatial embeddings \subref{fig:hypermodels:a}; in the recurrent network itself, changing the LSTM weights \subref{fig:hypermodels:b}; or after the LSTM entirely \subref{fig:hypermodels:c}. Hypernetwork ($h$) and model layers parameterized by it ($f$) highlighted in green.}
  \label{fig:hypermodels}
\end{figure*}
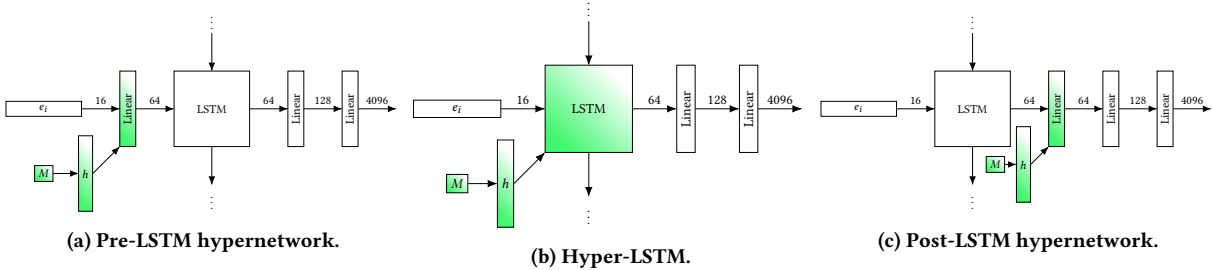

\subsection{GPS Output Layer}
\label{sec:output}

We follow a standard approach for this problem (including the original competition winner \cite{de2015artificial}): a softmax final layer generating a weight vector of a set of predefined points (which we call $\boldsymbol \alpha$). The final GPS vector is then a weighted average of the points,
\begin{equation}
    \hat{y} = \sum_j \alpha_j \tilde{p}_j
\end{equation}
Note that these points $\tilde{p}$ are the same reference points used in \cref{sec:geospatial_encoding} for the model's \emph{input}.  Thus $\tilde{p}$ are used to transform GPS points both to and from higher-dimensional embedding spaces used internally by the model, and $\boldsymbol \alpha$ serves a similar role to $\boldsymbol \delta$ (\cref{sec:geospatial_encoding}), both softmax weight vectors over these reference points.

Note that for every point $p_i \in T$, we have a predicted destination $\hat{y}_i$ predicted from $p_1, \ldots, p_{i}$. The training loss for a given trajectory $T$ is simply the mean Haversine distance between these predictions and the true destination $p_N$:
\begin{equation}
    \mathcal{L}(\hat{y}, p_N) = \frac{1}{N} \sum_{i=1}^N d_H(\hat{y}_i, p_N)
    \label{eq:loss}
\end{equation}
The evaluation metric is the Mean Haversine Distance (MHD), or the quantity in \cref{eq:loss} averaged over all trajectories in the validation set (a random sample of 10,000 trajectories held out from training). All models are trained for 10 epochs.

Following prior work, we will also consider the MHD over particular prefix lengths. For example, $MHD_{0.1}$ is the Haversine distance $d_H(\hat{y}_i, p_N)$ when $i$ corresponds to the first 10 percent of a trajectory, averaged over the validation set of trajectories.

\section{Experimental Results}

\begin{table*}[]
    \centering
    \begin{tabular}{lrrrrrrr}
    \toprule
        \textbf{Model} & $MHD$ & $MHD_{0.1}$ & $MHD_{0.3}$ & $MHD_{0.5}$ & $MHD_{0.7}$ & $MHD_{0.9}$  \\
    \midrule
    \emph{Results reported by prior work} & \\
    \midrule
        \citet{liao2021taxi}*$\dagger$ & 1.481 & 2.848 & 2.175 & 1.254 & \textbf{0.560} & 0.572 \\
        \citet{ebel2020destination}    & 1.460 &  2.71$\ddagger$   & 1.97$\ddagger$   & 1.26$\ddagger$   & 0.84$\ddagger$   & 0.69$\ddagger$ \\
        \citet{ebel2020destination}*   & 1.430 & 2.69$\ddagger$   & 1.93$\ddagger$   & 1.237 & 0.82$\ddagger$   & 0.66$\ddagger$ \\
    \midrule
    \emph{Our results} & \\
    \midrule
        pre-LSTM & 1.354 & 2.482 & 1.844 & 1.225 & 0.729 & 0.394 \\
        hyper-LSTM & \textit{1.320} & \textit{2.459} & \textit{1.825} & 1.214 & 0.691 & \textbf{0.334} \\
        \textbf{post-LSTM} & \textbf{1.317} & \textbf{2.429} & \textbf{1.800} & \textbf{1.195} & \textit{0.678} & \textit{0.335} \\
    \bottomrule
    \end{tabular}
    \caption{Our results relative to previously published ones. All results in kilometers (km). Best result in bold and second-best in italics. See \cref{sec:hypernet_comparison}. *: This result includes additional data added to the Porto dataset (e.g., weather \cite{ebel2020destination}, road network \cite{liao2021taxi}). $\dagger$: Included due to similar methods and results to \cite{ebel2020destination}, but note subtle differences in preprocessing. $\ddagger$: Approximated from published graphical results.}
    \label{tab:results}
\end{table*}

\begin{table}[]
    \centering
    \begin{tabular}{lr}
    \toprule
        \textbf{Model} & $MHD$ \\
    \midrule
        \textbf{Full model (post-LSTM, \cref{tab:results})} & \textbf{1.317} \\
    \midrule
        \emph{Remove hypernetwork} (\cref{sec:ablation:hyper}) & \\
    \midrule
        Concatenation & 1.432 \\
        No metadata & 1.382 \\
    \midrule
        \emph{Single timescales} (\cref{sec:ablation:time}) & \\
    \midrule
        Day only & 1.322 \\
        Week only & 1.329 \\
        Year only & 1.337 \\
    \bottomrule
    \end{tabular}
    \caption{Ablations of the best-performing hypernetwork. All results in kilometers (km). See \cref{sec:ablation}.}
    \label{tab:ablation}
\end{table}

\begin{figure}
    \centering
    \includegraphics[width=.45\textwidth]{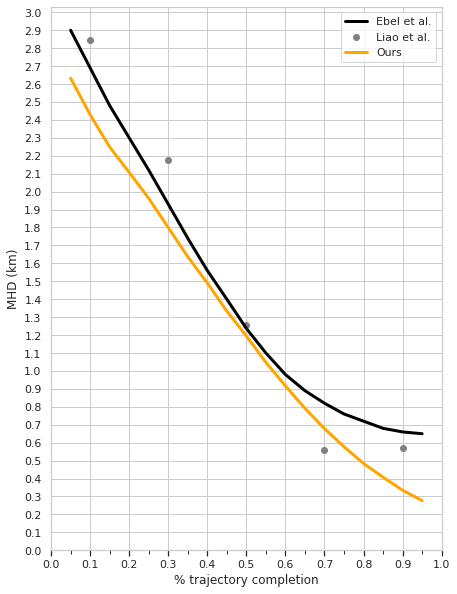}
    \caption{Comparison of the mean Haversine distance (MHD) of our best-performing model (post-LSTM; \cref{fig:hypermodels:c}) vs two strong published results (\citet{ebel2020destination} and \citet{liao2021taxi}). Lower MHD denotes better performance. See also \cref{tab:results}.}
    \label{fig:vs_sota}
\end{figure}

\begin{figure}
    \centering
    \includegraphics[width=.45\textwidth]{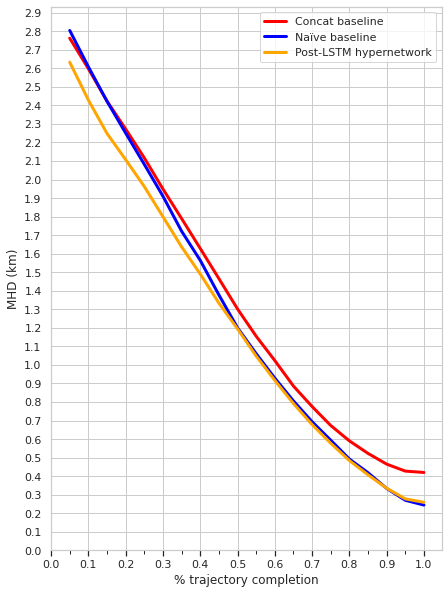}
    \caption{Ablation of our hypernetwork. The superior performance (lower MHD) of the hypernetwork (see \cref{tab:ablation}) seems especially concentrated in the first half of the trajectory. When most of the trajectory points have been provided, a simple LSTM without metadata is competitive; however, when few GPS points are provided, the hypernetwork significantly outperforms both alternatives.}
    \label{fig:vs_ablation}
\end{figure}

\subsection{Model comparison}
\label{sec:hypernet_comparison}

Despite the Porto dataset's original use as a standardized benchmark for open competition, design choices in subsequent work make cross-paper comparison difficult. Firstly, different papers often augment the dataset with their own metadata not present in the original release, which may give some models an advantage over others independent of architecture or training design.  For example, \citet{rossi2019modelling} incorporate point-of-interest (POI) data from Foursquare; \citet{liao2021taxi} include road networks from OpenStreetMap; \citet{ebel2020destination} train with (and without) additional weather data. Additionally, different teams have used different preprocessing pipelines, with potentially significant deviations in the resulting training (and validation) data.  Particularly problematic is the removal or alteration of ``difficult'' trajectories, such as those with GPS measurement error or destinations far outside Porto.  With roughly 1.7 million trajectories to begin, \citet{liao2021taxi} ``select the data located in the main areas of the city,'' resulting in 665,989 trajectories; \citet{ebel2020destination} delete trips outside the Porto area and maintain 1,545,240 trajectories.  Other rule-out criteria include trajectory length (e.g., \cite{ebel2020destination, zhang2018multi}) or missing GPS points, detected due to large instantaneous speed between two points (e.g., \cite{lam2015blue, ebel2020destination}).  Taken together, all these steps have the effect of removing the most difficult-to-predict trajectories from training and validation, biasing any evaluation against other papers' published metrics.

Given these limitations, \citet{ebel2020destination}'s result (without weather added) is, to our knowledge, the best published performance of any model without substantial additional metadata. We therefore take this as our baseline, and re-implement its preprocessing pipeline, validation split, etc. for fair comparison.  We note, however, that the more recent results of \citet{liao2021taxi} are numerically similar to those of \citet{ebel2020destination} despite the addition of a road network and more aggressive subsampling of the trajectories during preprocessing.  We outperform both methods, and both methods in turn outperform their authors' custom baselines and the original competition winner \cite{de2015artificial}.

Because hypernetworks have not been applied to this problem before, we investigate the optimal stage of fusing metadata (including time) with the recurrent model.  A hypernetwork can be used to parameterize the weights at any arbitrary layer of the model (\cref{fig:hypermodels}). We compare three possibilities (\cref{fig:hypermodels}):
\begin{enumerate}
    \item The hypernetwork $h$ parameterizes a linear layer placed \emph{before} the LSTM (\cref{fig:hypermodels:a}).  We call this the ``pre-LSTM'' model.
    \item The hypernetwork $h$ parameterizes all weights of the LSTM itself (\cref{fig:hypermodels:b}). We call this the ``hyper-LSTM'' model.
    \item The hypernetwork $h$ parameterizes a linear layer placed \emph{after} the LSTM (\cref{fig:hypermodels:c}). We call this the ``post-LSTM'' model.
\end{enumerate}We show the results, relative to \citet{ebel2020destination} and \citet{liao2021taxi}, in \cref{tab:results,fig:vs_sota}. 

Comparing the two prior publications, we see that despite similar $MHD$ values overall, \citet{ebel2020destination}'s method tends to perform better early in the trajectory and \citet{liao2021taxi} performs better late. This is perhaps because \citet{liao2021taxi} train only on prefixes of length 70\%, as opposed to varying lengths, allowing the model to specialize in late-trajectory prediction.  Note that \citet{liao2021taxi}'s error \emph{increases} when more of the trajectory, 90\%, is provided, suggesting the model's preference for 70\% prefixes.

Our proposed methods consistently outperform both methods in both the primary competition metric $MHD$ (\cref{tab:results}) and most prefix lengths individually (\cref{tab:results,fig:vs_sota}), with \citet{liao2021taxi} at 70\% the only exception.  In the first third of the trajectory, our method cleanly outperforms both prior papers. In the middle, all three methods are competitive, although ours remains narrowly better (by about 0.05 km). In the final third, \citet{ebel2020destination} and \citet{liao2021taxi} plateau around 0.66 and 0.56 km respectively, while our method continues to sharply reduce its error as more GPS points are provided, achieving errors below 0.3 km.

We also see that later fusion generally outperforms early fusion, and in particular placing the hypernetwork prior to the LSTM leads to weaker performance (\cref{tab:results}), while the other two methods (hyper- and post-LSTM) are competitive. However, all three hypernetworks outperform the prior state-of-the-art method, even without additional data such as weather conditions \cite{ebel2020destination} or a ground-truth road network \cite{liao2021taxi}.

\begin{figure*}
  \centering
    \includegraphics[height=0.35\textwidth]{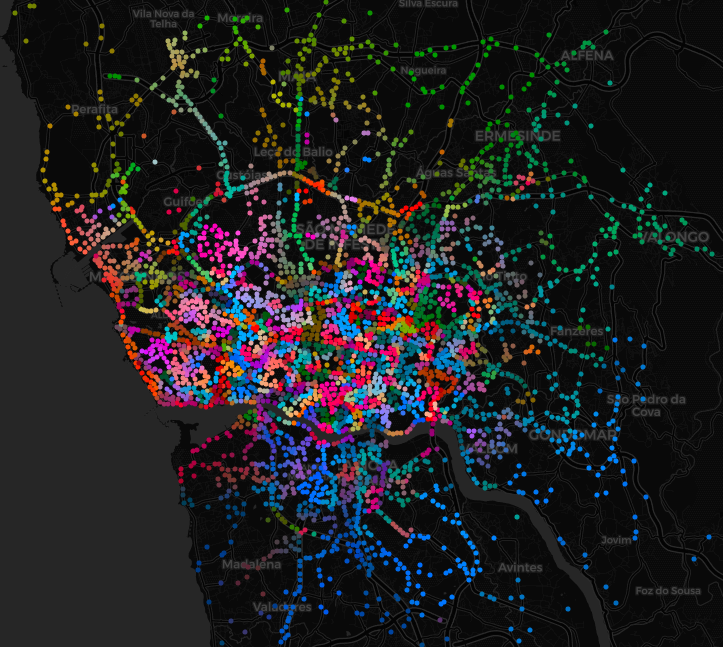}
    \includegraphics[height=0.35\textwidth]{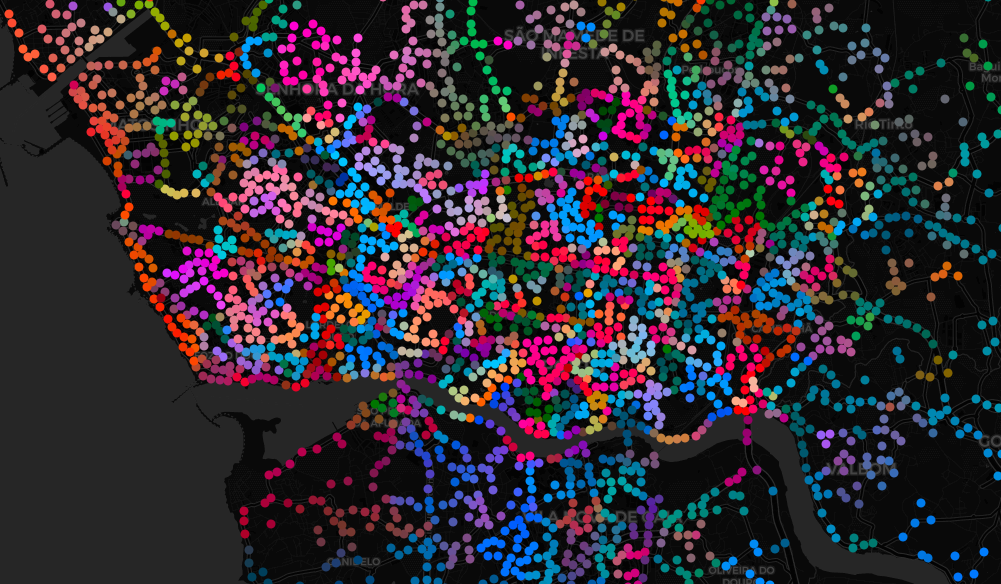}
    \caption{Visualizing embeddings of the Porto reference points. Left, the larger Porto area; right, focused on the downtown area. The model naturally learns an apparent unsupervised segmentation of the geospatial area with strong spatial locality. North of the city, highways are segmented into distinct colors near major turns or exits; in the west, beaches are a consistent color (reddish-orange) interrupted only by the historic site Castelo do Queijo (gray); downtown, the city appears to be grouped into distinct neighborhoods and districts with variable sizes and shapes and distinct borders. Best viewed in color.}
  \label{fig:embedding_visualization}
\end{figure*}

\subsection{Ablations}
\label{sec:ablation}
\subsubsection{Hypernetwork versus alternatives.}
\label{sec:ablation:hyper}
In addition to the numeric results reported by prior work, we also implement custom baselines which do not use hypernetworks.  This ensures that unique aspects of our approach, such as the geospatial encoding (\cref{sec:geospatial_encoding}) and model hyperparameters (e.g., layer sizes) remain constant, quantifying the exact contribution of the hypernetwork.  We consider two models:
\begin{enumerate}
    \item A model which concatenates the metadata with the main geospatial features after the LSTM (similar to \citet{ebel2020destination} and \citet{liao2021taxi}, among many others) then passes them through a fully-connected layer. We refer to this as the ``concatenation baseline.''
    \item A model which does not use the metadata at all. We refer to this as the ``na{\"i}ve baseline.''
\end{enumerate}
Both ablations underperform our main model (\cref{tab:ablation}). Removing the hypernetwork and replacing it with prior approaches degrades model performance.

\cref{fig:vs_ablation} further illustrates the circumstances under which our hypernetwork improves upon more common approaches. In the first half of the trajectory, the hypernetwork outperforms both the ablations by a significant margin (concatenation: 129 meters; na{\"i}ve: 171 meters). However, in the second half (when a majority of the trajectory points are visible to the model) the na{\"i}ve baseline becomes competitive: the geospatial information alone is sufficient for the prediction. Thus, the hypernetwork's improved performance draws from the most difficult cases, where few points have been provided and they are inadequate to predict the final destination alone, as we hypothesized in \cref{fig:teaser}.

\subsubsection{Different components of time.}
\label{sec:ablation:time}
Recall from \cref{sec:metadata} that we model time as sinusoids of period 1 day, 1 week, and 1 year. To evaluate each timescale's contribution, we modify the best-performing model (\cref{tab:results}) to produce three more models. Each ablated model receives only one timescale---either day, week, or year---in addition to the remaining metadata.

We can see in \cref{tab:ablation} that all three ablations perform slightly worse than the full model, but better than the concatenation or na{\"i}ve baselines. We therefore conclude that all three timescales---daily, weekly, and yearly patterns---contribute in some degree to the model's performance.

\subsection{Visualizing the region encodings}

Finally, we qualitatively investigate the geospatial encoding proposed in \cref{sec:geospatial_encoding}. Recall that each of the $N_{ref} = 4096$ reference points is associated with a learned embedding vector of size $m=16$. We employ t-SNE \cite{van2008visualizing} to reduce these embedding vectors into a 3D space, which we then rescale and interpret as color values. This allows us to map all embedding vectors to colors; we can then plot the GPS reference points (as originally shown in \cref{fig:geospatial_encoding}), but now color-coded by their embeddings to inspect them for structure.

The results of this visualization are shown in \cref{fig:embedding_visualization}. Compared to prior works that impose uniform grids, manual road network segmentations, $k$-d or quadtrees, etc., our results show a categorically better expression of Porto's geospatial structure. The t-SNE dimensionality reduction operates on the embedding vectors, \emph{not} the GPS values, but the embeddings still display strong geospatial locality. The embeddings appear to segment Porto into distinct learned regions. In the north outside the main city, where GPS points are sparse, highways are split into segments near major turns and exits.  In contrast, the denser downtown area appears grouped into neighborhoods and districts of variable size and shape. In a unique example, Porto's beaches (the western edge of the city) all take a similar reddish-orange embedding, with a small brownish-gray subregion around the Castelo do Queijo, a historic seaside fortress and major landmark. We can also see evidence in some locations of ``blending'' or interpolation for points between districts. 

This structure is learned entirely without additional supervision, simply as a means to achieve the model's primary task of destination prediction. Its adaptability to, and expressiveness of, Porto's geospatial structure cannot be matched by recently published state-of-the-art, which simply assigns input points to variably-sized rectangles from $k$-d trees \cite{ebel2020destination} or uniform grids and variably-sized squares from quadtrees \cite{liao2021taxi}.

\section{Conclusion}

We propose a model for destination prediction tasks which incorporates novel geospatial and temporal representations, and we validate them by achieving state-of-the-art performance on the Porto dataset. Our thorough ablation experiments confirm that our hypernetwork outperforms a concatenation-based approach common in prior work, and that different timescales each play a role in the improving the model's performance. We anticipate that improved prediction of vehicle destinations will be useful in urban planning, ride-sharing, and intelligent transportation applications.

\begin{acks}
This research was developed with funding from the Defense Advanced Research Projects Agency (DARPA). The views, opinions and/or findings expressed are those of the author and should not be interpreted as representing the official views or policies of the Department of Defense or the U.S. Government. Distribution Statement ``A'' (Approved for Public Release, Distribution Unlimited).
\end{acks}

\bibliographystyle{ACM-Reference-Format}
\bibliography{references}

\end{document}